\documentclass[runningheads]{llncs}

\usepackage{eccv}
\usepackage[margin=1.2in]{geometry}

\usepackage{eccvabbrv}
\usepackage[table]{xcolor}
\usepackage{graphicx}
\usepackage{booktabs}
\usepackage{comment}

\usepackage[accsupp]{axessibility}  

\usepackage{hyperref}
\usepackage{orcidlink}

\begin{document}

\title{EvoIQA - 
Explaining Image Distortions with Evolved White-Box Logic} 



\author{Ruchika Gupta\inst{1}\orcidlink{0009-0004-0696-2902} \and
Illya Bakurov\inst{2,3}\orcidlink{0000-0002-6458-942X} \and
Nathan Haut \inst{3}\orcidlink{0000-0002-3989-6532} \and
Wolfgang Banzhaf \inst{3}\orcidlink{0000-0002-6382-3245}}

\authorrunning{Gupta et al.}

\institute{Michigan State University \\
\email{guptaru1@msu.edu}}

\maketitle

\begin{abstract}
Traditional Image Quality Assessment (IQA) metrics typically fall into one of two extremes: rigid, hand-crafted mathematical models or "black-box" deep learning architectures that completely lack interpretability. To bridge this gap, we propose EvoIQA, a fully explainable symbolic regression framework based on Genetic Programming that \textbf{Evo}lves explicit, human-readable mathematical formulas for image quality assesment (\textbf{IQA}).
Utilizing a rich terminal set from the VSI, VIF, FSIM, and HaarPSI metrics, our framework inherently maps structural, chromatic, and information-theoretic degradations into observable mathematical equations. Our results demonstrate that the evolved GP models consistently achieve strong alignment between the predictions and human visual preferences. Furthermore, they not only outperform traditional hand-crafted metrics but also achieve performance parity with complex, state-of-the-art deep learning models like DB-CNN, proving that we no longer have to sacrifice interpretability for state-of-the-art performance.

\end{abstract}

\section{Introduction}
\label{sec:intro}

Digital images and videos are routinely compressed to meet bandwidth and storage constraints, inevitably introducing perceptible distortions \cite{nuzhat2021image, gu2020image}. To optimize these transmission systems, Image Quality Assessment (IQA) models aim to quantify these degradations by computationally mimicking the Human Visual System (HVS) \cite{wang2004image}.
IQA methods often belong to one of three categories: Full-Reference (FR) IQA \cite{ding2020image, ding2021locally, wang2003multiscale, wang2004image} which evaluates a distorted image by comparing it to a clean pristine reference image. Reduced Reference (RR) \cite{wang2011reduced, lv2009reduced} approaches possess some information about the reference image (e.g. watermark) but not the actual reference image and No-Reference (NR) IQA \cite{yang2022maniqa, mittal2012making} estimates the perceived quality of a distorted image without the presence of a reference image. Although all three paradigms serve critical roles in visual processing, this paper focuses specifically on FR-IQA. Existing methods fundamentally function as opaque "score generators." While they output a final quality score, they completely hide the human-like reasoning process of weighing distinct, multi-dimensional distortions (e.g., structural damage versus color shifts) \cite{you2024depicting}. State-of-the-art deep learning models capture these complex tradeoffs, but they trap this logic inside uninterpretable "black-box" architectures \cite{hassija2024interpreting}.
To better align IQA with transparent, human-like reasoning, we propose a new paradigm that moves beyond opaque score generation: Symbolic Regression (SR) driven by Genetic Programming (GP). Unlike traditional methods, SR dynamically searches for both the mathematical structure and the parameters required to explicitly model the observed data \cite{bnkf, schmidt2009distilling}. We drive this search using GP, an evolutionary algorithm that breeds potential solutions to efficiently navigate massive mathematical search spaces \cite{augusto2000symbolic, brameier2004linear}. These approaches are overwhelmingly preferred where interpretability and explicit mathematical relationships are desired \cite{makke2024interpretable, li2022symbolic}.

The main contributions of this paper can be summarized as follows:
\begin{itemize}
   \item \textbf{Statistical Feature Decomposition \& GP :} We extract statistical representations from existing IQA metrics to evolve a robust, mathematically explicit Genetic Programming (GP) model for IQA.

   \item \textbf{State-of-the-Art (SOTA) Symbolic Performance:} Our evolved $M_{\text{EvoIQ-Full}}$ model achieves a max SRCC of 0.8979 on TID2013 while outperforming specialized deep learning architectures like MEON and DB-CNN on 18/24 distortions using fewer than 10 unique parameters.

    \item \textbf{Explainable AI (XAI) for IQA:} We provide two transparent `White-Box' symbolic models by fusing Natural Scene Statistics (NSS) logic as detailed in \cref{sec:Full-metric}.
\end{itemize}

\section{Background}
\label{sec:background}
Image quality perception can be conducted through subjective or objective methods. Subjective quality is measured using the Mean Opinion Score (MOS) \cite{chubarau2020perceptual, saupe2016crowd} which averages ratings assigned by human observers. In contrast, the Difference of Mean Opinion Score (DMOS) quantifies the perceived quality degradation by comparing the MOS of a distorted image with that of its corresponding reference image. Because subjective testing is prohibitively expensive and difficult to scale for real-world applications \cite{streijl2016mean}, objective Image Quality Assessment (IQA) models have been developed to algorithmically predict these human evaluations. 
Historically, FR-IQA methods have evolved from simplistic pixel-wise error visibility metrics, such as Mean Squared Error (MSE) \cite{bovik2010handbook} and Peak Signal-to-Noise Ratio (PSNR) which is commonly applied to quantify the quality of image reconstruction and lossy compression \cite{hore2010image, ponomarenko2007between}. Recognizing the limitations of pixel-level differences, the field developed arguably the most well-known FR-IQA metric—the Structural Similarity (SSIM) index \cite{wang2004image}. SSIM evaluates image degradation based on luminance, contrast and structure, inspiring numerous multi-scale and wavelet-domain variants to better capture localized distortions \cite{wang2003multiscale, li2010content, liu2009wavelet, wang2010information, sun2018spsim, frackiewicz2021improved}.Concurrently, information-theoretic approaches like Visual Information Fidelity (VIF) \cite{sheikh2005information} emerged to quantify perceptual quality by measuring the mutual information extracted from Gaussian scale mixtures in the wavelet domain.

Recently, data-driven deep learning models, particularly Convolutional Neural Networks (CNNs), have achieved state-of-the-art accuracy by automatically learning complex relationships between image features and perceptual quality\cite{wu2020end, zhu2021generalizable, eybposh2024coniqa, bosse2017deep, ma2018end, zhang2018blind, tang2018full, ahn2021deep}. However, these networks inherently function as opaque "black boxes." To combine the strengths of various metrics without relying entirely on deep learning, fusion-based methods seek to improve predictive accuracy by combining the outputs of existing FR-IQA metrics. Researchers have explored a variety of techniques ranging from arithmetic optimization \cite{okarma2010combined} and linear/kernel ridge regression \cite{ oszust2017image, yuan2015image, liu2012image}.

Within this fusion paradigm, evolutionary algorithms have been employed to optimize metric performance \cite{oszust2015decision, merzougui2021multi, correia2022towards}. For example, Genetic Algorithms (GAs) have been successfully employed to fine-tune the internal scalar parameters and exponents of established equations like SSIM to better align with subjective evaluations\cite{bakurov2020parameters, bakurov2022genetic, bakurov2022structural}. Recently, Bakurov et al. \cite{bakurov2023full} introduced a GP framework to evolve SSIM-like quality measures from fundamental image statistics. In this paper, we significantly expand upon this paradigm by introducing EvoIQA, which utilizes a vastly richer terminal set and higher-order statistics detailed in the following section.

\section{Proposed Methodology}
\label{sec:method}
We propose Evo-IQA, a framework that decomposes established IQA metrics into their constituent perceptual components. Using the PIQ library \cite{kastryulin2022piq} as our foundational implementation layer, we extract low-level feature maps and characterize them using the Asymmetric Generalized Gaussian Distribution (AGGD) \cite{lasmar2009multiscale}. We leverage a two-phased Stack-based Genetic Programming (StackGP) \cite{stackGPDocs} architecture to evolve two distinct solutions:

\begin{itemize}
    \item \textbf{EvoIQA-Full:} Utilizes the complete terminal set to maximize predictive accuracy.
    \item \textbf{EvoIQA-Subset:} Employs a reduced feature set—selected via Random Forest importance—to optimize computational efficiency.
\end{itemize}

\subsection{The Low-Level Feature Decomposition}
\label{sec:features}
For each IQA metric $i$ implemented in \cite{kastryulin2022piq}, we compute $K$ distinct perceptual maps. We define the spatial extraction of the $k$-th map (e.g., Phase Congruency or Gradient Magnitude) as:

\begin{equation}
M_{i, k} = \mathcal{E}_{i, k}(I_{ref}, I_{dist})
\end{equation}

To translate these high-dimensional spatial maps into a compact feature representation for symbolic evolution, we treat the coefficients of $M_{i, k}$ as a stochastic distribution. Each map is parameterized by its AGGD descriptor vector $\Theta_{i, k}$ which is further explained in \cref{sec:aggd_theory}, forming our terminal set $\mathcal{T}$:

\begin{equation}
\Theta_{i, k} = { \hat{\alpha}, \bar{\sigma}^2, \mu, \sigma, \kappa, \gamma }
\label{eq:aggd_breakdown}
\end{equation}

These higher-order statistics capture the "texture of the error" \cite{yao2021no},  allowing the model to remain sensitive to localized artifacts (e.g., JPEG blockiness) that global mean-based metrics often overlook as shown in \cref{fig:four_map_comparison}. The StackGP architecture then evolves a symbolic mapping function $f: \mathcal{T} \rightarrow \mathbb{R}$, effectively discovering the non-linear mathematical equation that best correlates these statistical features with human perceptual scores:

\begin{equation}
\widehat{MOS} = f(\Theta_{1,1}, \dots, \Theta_{i,k})
\end{equation}

The following subsections provide a detailed breakdown of the constituent perceptual metrics.

\subsubsection{HaarPSI}

Following the HaarPSI framework \cite{kastryulin2022piq, reisenhofer2018haar}, the reference and distorted images ($x$ and $y$) are transformed into the YIQ color space \cite{lakhwani2015color}. The luminance component $x_{Y}$ is analyzed using a multiscale Haar wavelet decomposition. For each scale $s \in \{1, 2, 3\}$, the Haar coefficient maps are computed by convolving the image with horizontal ($h_1$), vertical ($h_2$), and diagonal ($h_3$) kernels:

\begin{equation}
    C_{x, k}^{(s)} = (x_{Y} \ast h_{k}^{(s)}) \downarrow_{2^{s-1}}, \quad k \in \{1, 2, 3\}
\end{equation}

Following the HVS-based logic of the HaarPSI model \cite{reisenhofer2018haar}, a weight map $W$ is extracted from the last scale ($s=3$) to capture the low-frequency structural backbone of the image. The weights are computed as follows:

\begin{equation}W = \max_{k \in {1, 2, 3}} \left( \left| C_{x, k}^{(3)} \right|, \left| C_{y, k}^{(3)} \right| \right)\end{equation}

For color images, the feature extraction framework is extended to the chrominance space (the $I$ and $Q$ channels of the YIQ color space). To emulate the low-pass filtering characteristics of the Human Visual System (HVS), we extract the absolute chrominance coefficients and apply a $2 \times 2$ average pooling operation. While traditional methods rely on local similarity mapping shown in \cref{eq:ssim_pc} \cite{wang2004image}, we use a statistical approach to quantify perceptual degradation by measuring the shift between the estimated probability density functions (PDFs) of the reference ($p_{x,c}$) and distorted ($p_{y,c}$) chrominance channel ($c \in \{I, Q\}$) . This perceptual distance is computed using the Kullback-Leibler (KL) Divergence \cite{verdicchio2008image}: 
\begin{equation}
D_{KL}(p_{x,c} \parallel p_{y,c}) = \sum_{z \in \mathcal{Z}} p_{x,c}(z) \log \left( \frac{p_{x,c}(z)}{p_{y,c}(z)} \right)
\label{eq:kl_chroma}
\end{equation}

\subsubsection{FSIM}
The core premise of FSIM is that the HVS perceives features at points where Fourier components are at their maximum phase, a phenomenon captured by Phase Congruency (PC) \cite{zhang2011fsim}. Phase Congruency is a dimensionless measure that is invariant to image contrast, making it highly robust to illumination changes. The reference ($r$) and distorted($d$) images are also converted to the YIQ color space \cite{lakhwani2015color}, where the Y-channel (luminance) is processed through a bank of Log-Gabor filters at multiple scales and orientations. The Phase Congruency map $PC(x)$ is calculated as the following ratio:\begin{equation}PC(x) = \frac{E(x)}{\sum_n A_n(x) + \epsilon}\end{equation}where $E(x)$ is the local energy and $A_n(x)$ represents the amplitude of the $n$-th Fourier component. The similarity between the reference and distorted phase congruency maps, denoted as $S_{PC}$, is computed using a stability-constrained ratio \cite{wang2004image}:\begin{equation}S_{PC}(x) = \frac{2 \cdot PC_r(x) \cdot PC_d(x) + T_1}{PC_r^2(x) + PC_d^2(x) + T_1} \label{eq:ssim_pc} \end{equation}where $T_1$ is a fixed positive constant ($T_1 = 0.85$). 
To prioritize regions of high structural significance, a pixel-wise weight map is derived by taking the maximum of the reference and distorted PC maps, $PC_{map}(x) = \max(PC_r(x), PC_d(x))$. These maps are then processed into the statistical vector $\mathcal{T}_i$ to characterize the distribution of structural degradation. 

\subsubsection{VIF}
To quantify visual information loss, we incorporate components of the Visual Information Fidelity (VIF) index \cite{sheikh2006image, kastryulin2022piq}. VIF models the distortion process as an information leakage. Focusing on the final scale ($s=4$) to capture coarse structural information, the relationship between a local reference neighborhood ($x$) and its distorted counterpart ($y$) is modeled in \cref{eq:vif_model} where $v$ is additive white Gaussian noise and $g$ is the the deterministic gain representing preserved signal strength. From the gain map ($g$) we then extract the six global statistical aggregators shown in \cref{eq:aggd_breakdown}

 \begin{equation}
    y = g \cdot x + v, \quad \text{where} \quad g = \frac{\sigma_{xy} }{\sigma_{x}^2 + \epsilon}
    \label{eq:vif_model}
\end{equation}

\subsubsection{VSI}
To emulate the Human Visual System (HVS), we extract Visual Saliency-Induced (VSI) features \cite{zhang2014vsi}. Images are transformed into the $LMN$ color space, downsampled \cite{zhang2014vsi}, and processed via a Scharr operator on the luminance channel to calculate Gradient Magnitude ($GM$) \cite{kastryulin2022piq}. Structural and chromatic similarity maps ($S_{GM}$, $S_{MN}$) are then computed using the stability-constrained ratio in \cref{eq:vif_model}. To optimize the Genetic Programming terminal set, the chromatic maps ($S_M$, and $S_N$) are averaged.

\subsection{AGGD Representation of Low-Level Features}
\label{sec:aggd_theory}
Our characterization builds on the Natural Scene Statistics (NSS) paradigm introduced by BRISQUE \cite{mittal2012no}, which established that the Mean Subtracted Contrast Normalized (MSCN) coefficients of pristine images follow a predictable Gaussian distribution. While distortions induce measurable shifts in these statistics \cite{mittal2012no, sun2021blind}, the symmetric Gaussian model is often too restrictive to capture complex degradation. Consequently, we follow the approach in \cite{mittal2012no} by adopting the Asymmetric Generalized Gaussian Distribution (AGGD) \cite{lasmar2009multiscale, sharifi1995estimation}. Unlike the MSCN-based approach in BRISQUE, we extend this paradigm by applying AGGD fitting directly to perceptually-weighted similarity and spectral magnitude maps. This allows our model to quantify the 'statistical naturalness' of structural features rather than mere pixel-level fluctuations.
For a given feature map $M$, we account for potential statistical asymmetry introduced by non-linear distortions by partitioning the coefficients into negative ($M^{-}$) and positive ($M^{+}$) sets \cite{lasmar2009multiscale}. We estimate the left and right scale parameters, $\sigma_l$ and $\sigma_r$, using the Root Mean Square (RMS) of each partition:\begin{equation}\sigma_l = \sqrt{\frac{1}{|M^{-}|} \sum_{x \in M^{-}} x^2}, \quad \sigma_r = \sqrt{\frac{1}{|M^{+}|} \sum_{x \in M^{+}} x^2}\end{equation}

The total estimated scale parameter $\bar{\sigma}^2$, which serves as a variance-based feature for our Genetic Programming engine, is computed as the mean of $\sigma_l$ and $\sigma_r$.

The shape parameter $\alpha$ defines the shape of the distribution. Following the BRISQUE moment-matching procedure \cite{mittal2012no}, we first compute the empirical ratio of moments $\rho_{ratio}$ from the map $M$:
\begin{equation}
\rho_{ratio} = \frac{\left(\mathbb{E}[|M|]\right)^2}{\mathbb{E}[M^2]}
\end{equation}

The optimal shape parameter $\hat{\alpha}$ is found by solving for the minimum of the squared difference between the observed ratio and the theoretical AGGD moment ratio \cite{lasmar2009multiscale} over the interval $[0.2, 10]$:\begin{equation}\hat{\alpha} = \arg\min_{\alpha} \left[ \frac{\Gamma(2/\alpha)^2}{\Gamma(1/\alpha)\Gamma(3/\alpha)} - \rho_{ratio} \right]^2\end{equation}

To capture the heavy-tailed nature of localized artifacts, we supplement the AGGD parameters with higher-order descriptors—kurtosis ($\kappa$) and skewness ($\gamma$) \cite{li2020blind}:
\begin{equation}\kappa = \frac{\mathbb{E}[(M - \mu)^4]}{\sigma^4}, \quad \gamma = \frac{\mathbb{E}[(M - \mu)^3]}{\sigma^3}\end{equation}
These parameters—comprising the AGGD shape $\hat{\alpha}$ and scale $\bar{\sigma}^2$, along with empirical moments $\mu, \sigma, \kappa,$ and $\gamma$—constitute the final terminal vector $\Theta_{i, k}$ for a given metric $i$ and perceptual channel $k$. This hybrid representation ensures that both the global statistical distribution and the presence of localized, heavy-tailed artifacts are captured.

\begin{figure*}[htbp] 
    \centering

    \begin{subfigure}[b]{0.32\textwidth}
        \centering
        
        \includegraphics[width=\textwidth, height=2cm]{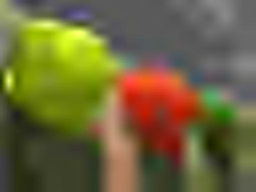}
        \caption{Highly Distorted Image}
    \end{subfigure}
    \hfill
    \begin{subfigure}[b]{0.32\textwidth}
        \centering
        \includegraphics[width=\textwidth, height=2cm]{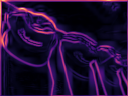}
        \caption{Level 5: Weight Map}
    \end{subfigure}
    \hfill
    \begin{subfigure}[b]{0.32\textwidth}
        \centering
        \includegraphics[width=\textwidth, height=2.3cm]{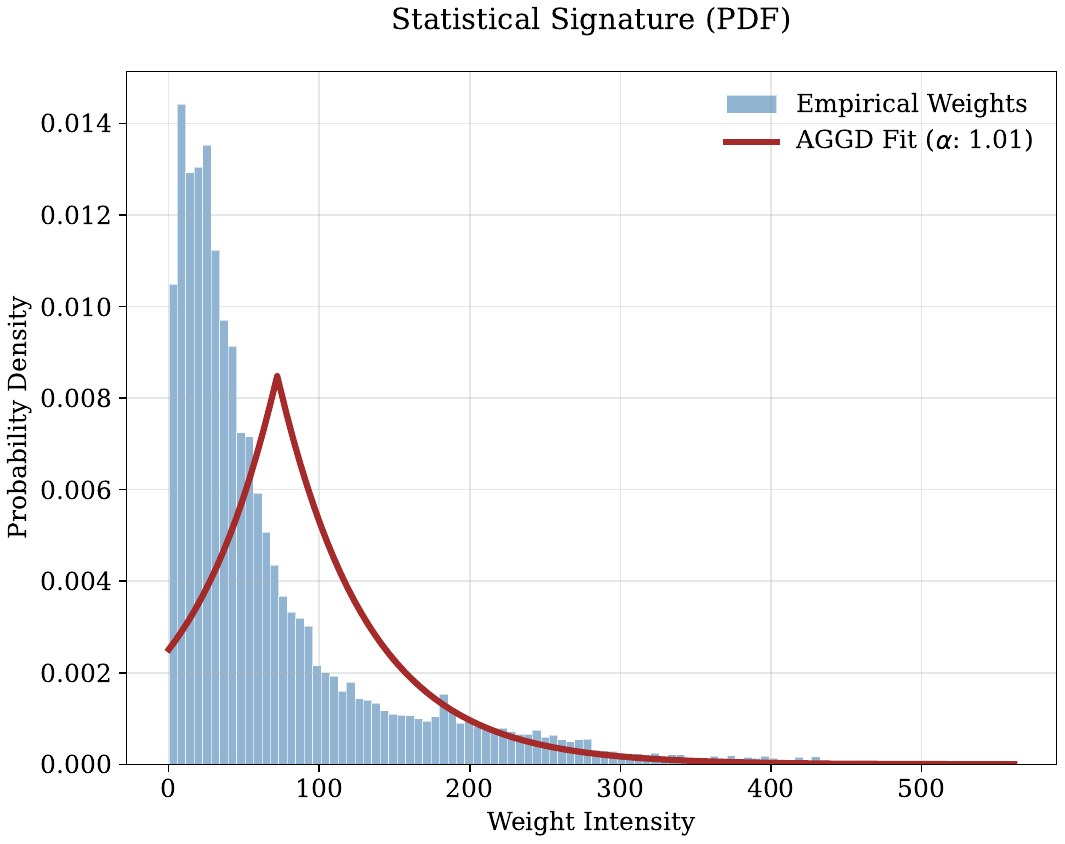}
        \caption{Level 5: AGGD Curve}
    \end{subfigure}

    \vspace{0.1cm} 

    \begin{subfigure}[b]{0.32\textwidth}
        \centering
        \includegraphics[width=\textwidth, height=2cm]{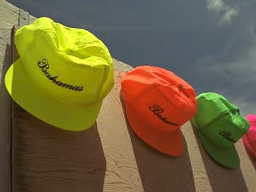}
        \caption{Less Distorted Image}
    \end{subfigure}
    \hfill
    \begin{subfigure}[b]{0.32\textwidth}
        \centering
        \includegraphics[width=\textwidth, height=2cm]{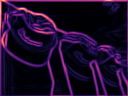}
        \caption{Level 2: Weight Map}
    \end{subfigure}
    \hfill
    \begin{subfigure}[b]{0.32\textwidth}
        \centering
        \includegraphics[width=\textwidth, height=2.3cm]{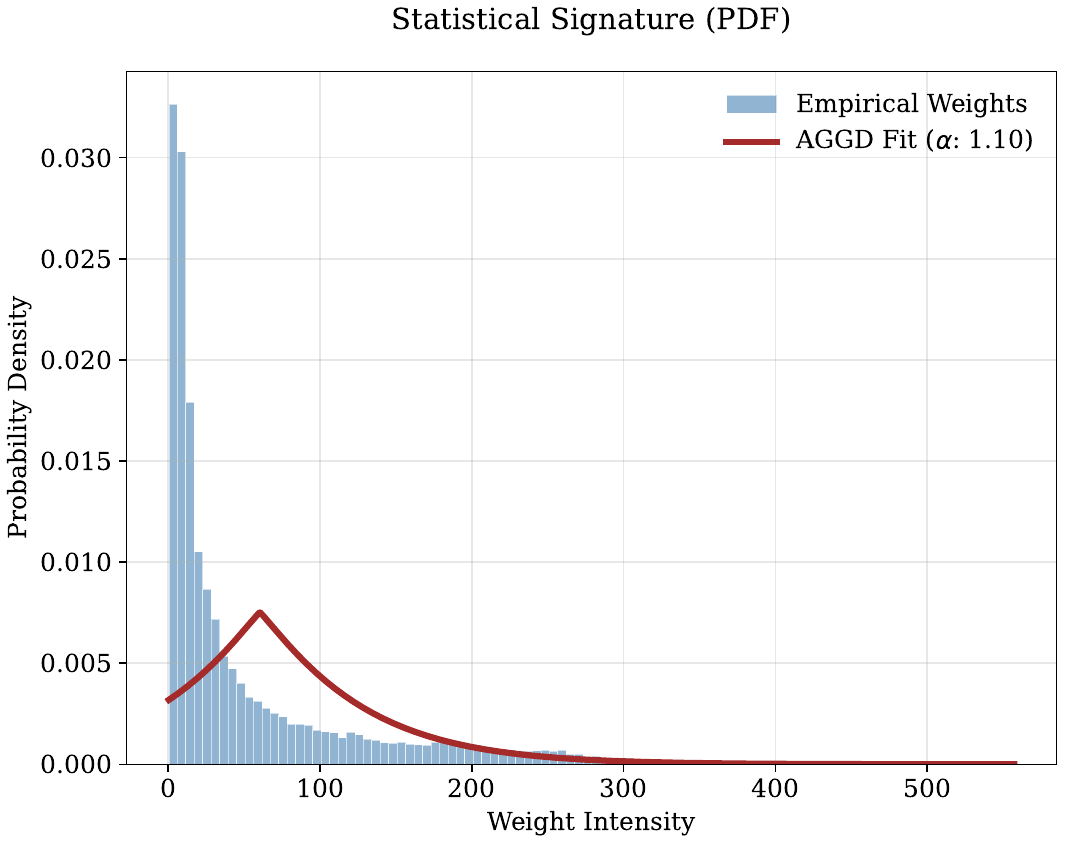}
        \caption{Level 2: AGGD Curve}
    \end{subfigure}
    
    \caption{Hierarchical Decomposition of JPEG2000 (JP2K) Compression. Each row illustrates the transition from visual artifacts to spatial weight maps and their corresponding AGGD signatures. Note the heavy compression in Row 1 results in a sharper peaked distribution.}
    \label{fig:full_width_comparison}
\end{figure*}

In our framework, the transition from spatial weight maps to AGGD signatures serves as a critical dimensionality reduction step for the symbolic regression engine. As shown in \cref{fig:full_width_comparison}, while the raw HaarPSI weight maps (Fig. 1b and 1e) capture the spatial layout of artifacts, they are often dominated by scene geometry. In contrast, the AGGD curves (Fig. 1c and 1f) effectively isolate the 'Statistical Peaking' caused by heavy compression. Specifically, the highly distorted image in Row 1 produces a significantly sharper distribution (lower $\hat{\alpha}$, higher $\kappa$). By providing the GP engine with these low-dimensional summaries rather than high-dimensional spatial maps, we significantly reduce the search space, enabling the evolution of more robust and interpretable quality equations.

\subsection{StackGP}\label{sec:stackgp}
StackGP is a stack-based genetic programming system implemented in Python that has primarily been applied to symbolic regression tasks \cite{haut2024active}. It implements several state of the art techniques, such as the multi-objective Pareto tournament selection \cite{pareto}, and a correlation fitness function \cite{Haut2023}. The search space is explored using three primary operators: cloning, two-point crossover for recombination, and mutation (comprising point-wise changes and structural stack modifications such as pushing, trimming, or random terminal/function insertion).

\textbf{Model Selection} Model selection is performed through Pareto Tournaments \cite{pareto}. In each tournament, five models are picked at random, and those that are "not dominated" are considered winners. A model is considered non-dominated if no other model in that group is both more accurate and simpler at the same time.

\textbf{Interpretability Constraints}
\label{subsec:gp_constraints}
To prevent the model from generating "spaghetti code," we enforce a restricted operator set $\mathcal{O}_{strict}$ to hold only arithmetic operations. We explicitly ban non-linear activation functions (e.g., $\exp, \sin $) from the evolutionary process.
\begin{equation}
\mathcal{O}{strict} = { +, -, \times,  \div{protected} }
\end{equation}
This constraint forces the GP to combine the computed features linearly. The impact of using a non-linear operator set can be found in \cref{tab:ablation}.

\textbf{Evolutionary Process}
\label{subsec:evolution_process}
We adopt a two-phase evolutionary strategy to allow for adequate exploration:

\begin{enumerate} \item \textbf{Warm-start Initialization:} We began with a preliminary "sprint" of 10 generations with a population of 500 individuals. During this stage, mutation was the sole variation operator used to foster exploration and rapidly filter out degenerate programs (e.g., constant or non-responsive functions). The surviving population from this sprint is preserved and passed as the \textit{initial seed} for the next stage, ensuring the main search begins from a baseline of viable candidates.

\item \textbf{Exploration Phase:} The main search phase utilized a larger population ($N=600$) which was evolved for 400 generations. Variation was driven by crossover (80\%) and mutation (20\%), utilizing a tournament selection of size 10. To prevent the loss of high-performing symbolic structures, an elitism rate of 15 was maintained. All other parameters for the StackGP configuration can be found here \cite{stackGPDocs}. 

\end{enumerate}
\section{Results}
\label{sec:results}
This section presents and discusses the experimental results. The analysis is based on Spearman's Rank-Order Correlation Coefficient (SROCC) which is a widely accepted evaluation measure for IQA metrics \cite{wang2004image, streijl2016mean}. \subsection{Experimental Setup}
We utilize TID2013 \cite{ponomarenko2015image} as our primary training corpus. To ensure statistical rigor, we conducted 30 independent runs with unique seeds using reference-level partitioning. For each run, 20\% of the reference images were reserved for a hold-out test set, while the remaining 80\% were split into training (80\%) and validation (20\%) subsets. This approach ensures the model is evaluated on entirely unseen visual content. To evaluate generalization, we validate the framework across five diverse IQA benchmarks:
\begin{itemize}
    \item \textbf{TID2013\cite{ponomarenko2015image} \& KADID-10k \cite{kadid10k}:} Large-scale corpora featuring 3,000 and 10,125 images respectively, across 24--25 distortion types. These cover a broad range from lab-controlled synthetic artifacts to crowdsourced digital art.
    \item \textbf{CSIQ \cite{larson2010most}:} Comprises 866 images across 6 primary distortions, focusing on natural scene photography from public-domain sources, predominantly from the United States National park service.
    \item \textbf{CID:IQ\cite{liu2014cid} \& VDID2014 \cite{gu2015quality}:} Specialized, viewing-distance-aware datasets containing 690 and 160 images respectively, designed to evaluate the impact of resolution and observer distance on perceived quality.
\end{itemize}

\subsection{Overall performance}
We evaluated two configurations: EvoIQA-Full (incorporating the exhaustive features detailed in \cref{sec:features}) and EvoIQA-Subset. The GP engine acted as an inherent regularizer, yielding simple, interpretable expressions that converged on a shared core of Scharr-filtered gradient maps ($\sigma_{gm}$), chrominance similarity ($\mu_{S_{mn}}$), and reference gradient magnitude ($\nabla_{ref}$). While the Random Forest ensemble achieved the highest numerical ceiling (SROCC 0.9390), its 'Black-Box' nature lacks the perceptual interpretability afforded by our symbolic approach. Furthermore, both EvoIQA variants consistently outperformed the SVR and RF baselines as shown in \cref{tab:comparison_results}, as well as the DNN architectures discussed in \cref{sec:dist_groups}. 

\subsection{The White-Box Advantage}
In the following sections, we select the top-performing models from the Pareto Front \cite{pareto} to analyze the perceptual significance of their symbolic structures.
\subsubsection{The EvoIQA-Full Metric}
\label{sec:Full-metric}
\begin{figure*}[t!]

    \centering
    \small
    \begin{minipage}[c]{0.18\textwidth}
        \centering
        \textbf{Reference}\\
        \vspace{0.1cm}
        \includegraphics[width=\textwidth]{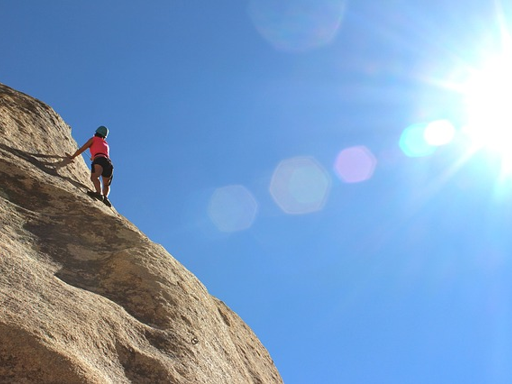}
    \end{minipage}
    \hfill
    \begin{minipage}[c]{0.78\textwidth}
        \centering
        \begin{tabular}{c@{\hspace{0.15cm}}c@{\hspace{0.15cm}}c@{\hspace{0.15cm}}c@{\hspace{0.15cm}}c}
            \includegraphics[width=0.18\textwidth]{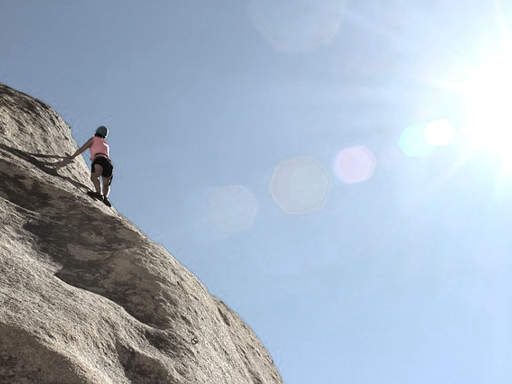} & 
            \includegraphics[width=0.18\textwidth]{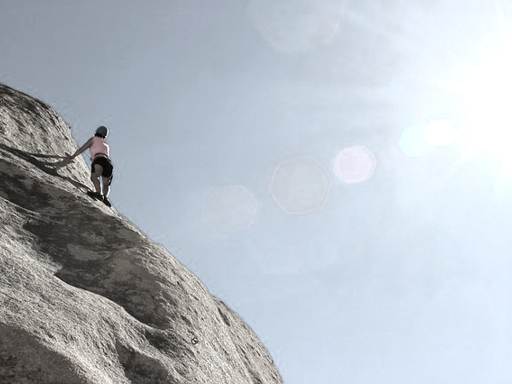} & 
            \includegraphics[width=0.18\textwidth]{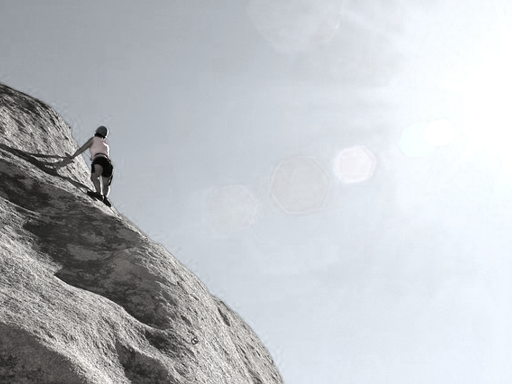} & 
            \includegraphics[width=0.18\textwidth]{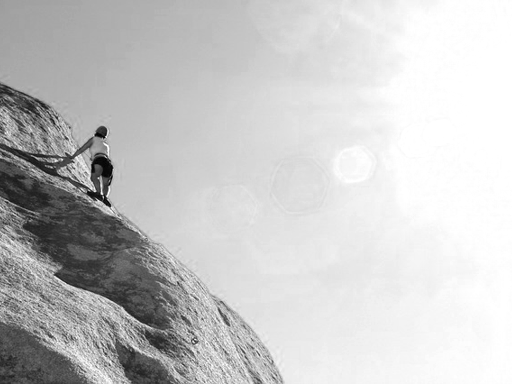} & 
            \includegraphics[width=0.18\textwidth]{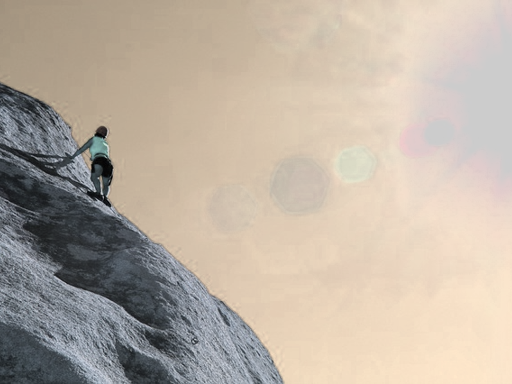} \\
            
            \scriptsize Dist. 7-1 & \scriptsize Dist. 7-2 & \scriptsize Dist. 7-3 & \scriptsize Dist. 7-4 & \scriptsize Dist. 7-5 \\
            \vspace{0.1cm} \\

            \includegraphics[width=0.18\textwidth]{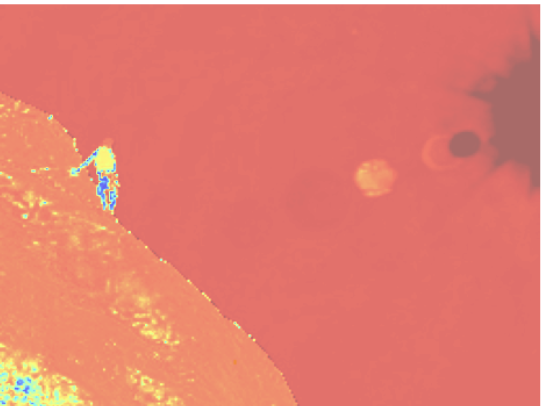} & 
            \includegraphics[width=0.18\textwidth]{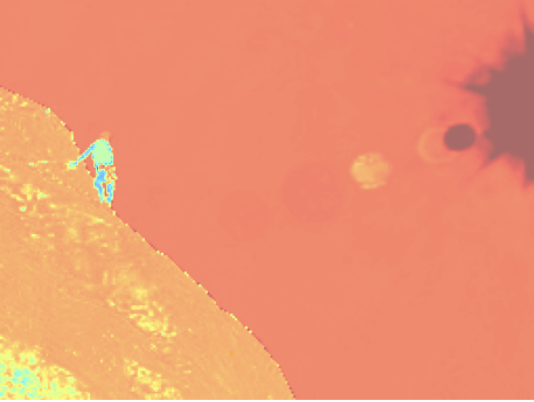} & 
            \includegraphics[width=0.18\textwidth]{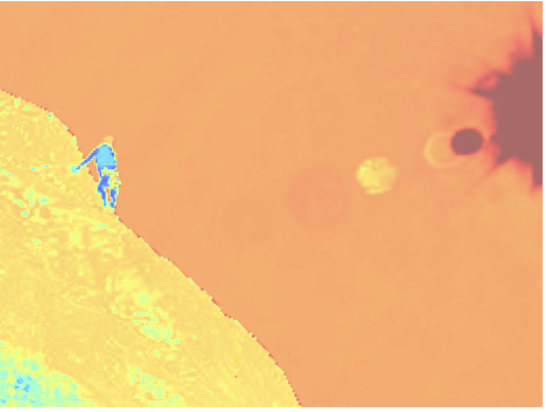} & 
            \includegraphics[width=0.18\textwidth]{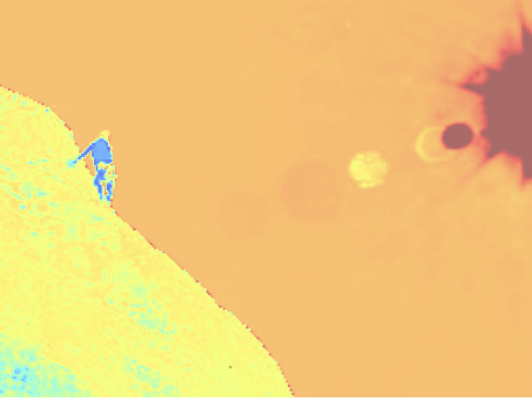} & 
            \includegraphics[width=0.18\textwidth]{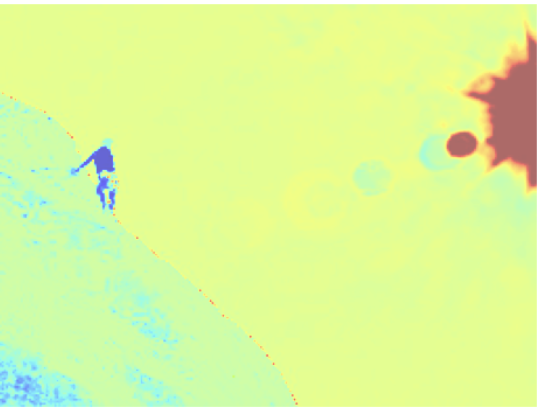} \\
            
            \scriptsize \textbf{$S_{mn}$ Map} & \scriptsize \textbf{$S_{mn}$ Map} & \scriptsize \textbf{$S_{mn}$ Map} & \scriptsize \textbf{$S_{mn}$ Map} & \scriptsize \textbf{$S_{mn}$ Map} \\
            \vspace{0.2cm} \\

            \scriptsize MOS: 2.97 & \scriptsize MOS: 2.13 & \scriptsize MOS: 2.03 & \scriptsize MOS: 2.03 & \scriptsize MOS: 1.73 \\
            \scriptsize Haar: 0.799 & \scriptsize Haar: 0.585 & \scriptsize Haar: 0.449 & \scriptsize Haar: 0.317 & \scriptsize \textcolor{red}{Haar: 0.679} \\
            \scriptsize \textbf{Ours: 3.91} & \scriptsize \textbf{Ours: 3.31} & \scriptsize \textbf{Ours: 3.01} & \scriptsize \textbf{Ours: 2.70} & \scriptsize \textbf{Ours: 1.80} \\
        \end{tabular}
    \end{minipage}

    \caption{\textbf{Qualitative comparison of color saturation artifacts in an unseen KADID-10k dataset.} The reference image (left) is compared against five increasing levels of saturation distortion. While the structural HaarPSI baseline remains invariant to the last chromatic shift, our \textbf{EvoIQA models leverage a learned feature subset}—specifically the evolved Chrominance Similarity ($S_{mn}$) maps—to successfully track perceptual degradation. The resulting predictions show a high monotonic alignment with human Mean Opinion Scores (MOS), whereas the baseline results (highlighted in \textcolor{red}{red}) fail to capture the intensity of the distortion.}
    \label{fig:qualitative_inline}
\end{figure*}

\begin{figure*}[t]
    \centering
    \begin{subfigure}[b]{0.48\textwidth}
        \centering
        \includegraphics{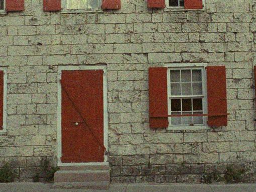}
        \caption{Global Additive Noise (I01\_01\_4)}
    \end{subfigure}
    \hfill
    \begin{subfigure}[b]{0.48\textwidth}
        \centering
        \includegraphics{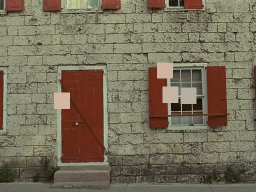}
        \caption{Local Block Distortion (I01\_15\_4)}
    \end{subfigure}

    \vspace{0.2 em}

    \begin{subfigure}[t]{0.24\textwidth}
        \centering
        \includegraphics[width=\textwidth]{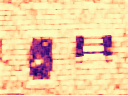}
        \caption{HaarPSI: Noise}
    \end{subfigure}
    \hfill
    \begin{subfigure}[t]{0.24\textwidth}
        \centering
        \includegraphics[width=\textwidth]{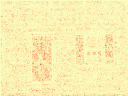}
        \caption{VSI Gradient Map \\ \small $\mu_g=0.97, \sigma_g=0.05$}
    \end{subfigure}
    \hfill
    \begin{subfigure}[t]{0.24\textwidth}
        \centering
        \includegraphics[width=\textwidth]{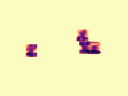}
        \caption{HaarPSI: Blocks}
    \end{subfigure}
    \hfill
    \begin{subfigure}[t]{0.24\textwidth}
        \centering
        \includegraphics[width=\textwidth]{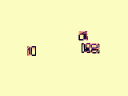}
        \caption{VSI Gradient Map \\ \small $\mu_g=0.98, \sigma_g=0.11$}
    \end{subfigure}

    \caption{\textbf{Localized vs. Global Distortion Analysis.} 
    Comparison of global noise (a) and local block distortion (b). While HaarPSI (c, e) blurs artifacts through multi-resolution pooling, our VSI Gradient map ($s_{gm}$) explicitly isolates unnatural edge discontinuities (d, f). Under identical means ($\mu \approx 0.98$), localized blocks trigger a $2\times$ spike in standard deviation ($\sigma = 0.11$ vs. $0.05$). This Gradient Heterogeneity allows our evolved GP model (Eq. \ref{eq:evoiq-subset}) to penalize structural collapse via the Coefficient of Variation ($\Omega_{cv} = \sigma_{gm}/\mu_{gm}$), identifying distortions that traditional pooling obscures.}
    \label{fig:four_map_comparison}
\end{figure*}

 The evolutionary process for the exhaustive feature set converged on a 7-feature symbolic expression. The discovered expression for $\mathcal{M}_{\text{EvoIQA-Full}}$ manifests a hierarchical structure.  As shown in \cref{fig:qualitative_inline} and detailed in \cref{tab:variables}, the term $(\sqrt{\mu_{\Delta C}} + \mu_{S_{mn}})$ represents a cross-space chromatic integrator, where the model simultaneously assesses color naturalness modulated by the reference gradient $\nabla_{ref}$ to account for foveal sensitivity in high-frequency regions. The global multiplication by $\sigma_{S_{gm}}$ (VSI Gradient Map) acts as a Contrast Masking gate, ensuring that structural and chromatic errors are only weighted in the most salient regions.
 The resulting metric, $\mathcal{M}_{\text{EvoIQA-Full}}$, is defined as:

\begin{equation}
Q = \sigma_{S_{gm}} \cdot \left[ \bar{\sigma}^2_{S_{gm}} + \nabla_{ref} \left( \sqrt{\mu_{\Delta C}} + \mu_{S_{mn}} - \mu_{PC_{max}} \right) + \sqrt{(\sigma_{S_{mn}})^2 + \sigma_{S_{gm}}} \right] + \sigma_{S_{gm}}
\label{eq:evoiq_full}
\end{equation}

\begin{table}[tb]
  \caption{Definitions of the perceptual primitives in the $\mathcal{M}_{\text{EvoIQA-Full}}$ metric.}
  \label{tab:variables}
  \centering
  \scriptsize
\renewcommand{\arraystretch}{1.2} 
  \setlength{\tabcolsep}{4pt} 
  \begin{tabular}{@{}ll p{6cm}@{}}
    \toprule
    Symbol & Feature Name & Perceptual Significance \\
    \midrule
    $\sigma_{gm}, \bar{\sigma}^2_{S_{gm}}$ & Gradient Fidelity & Scharr-filtered similarity maps, quantifying total distortion energy and local edge consistency. \\
    $\mu_{S_{mn}}, \sigma_{S_{mn}}$ & Chromatic Integrity & LMN-space chromatic similarity, tracking global color shifts and local artifacts to isolate perceived color degradation from luminance fluctuations. \\
    $\nabla_{ref}$ & Ref. Gradient & Gradient magnitude of the reference image, acting as a structural weighting factor. \\
    $\mu_{\Delta C}$ & Chromatic NSS & Mean $D_{KL}$ of AGGD-fitted PDFs derived from $2\times2$ average-pooled $I$ and $Q$ chrominance channels, quantifying statistical chromatic shift.  \\
    $\mu_{PC_{max}}$  & PC Saliency & Maximum phase congruency to identify salient regions - $\max(PC_r, PC_d)$. \\
    \bottomrule
  \end{tabular}
\end{table}

\subsubsection{The EvoIQA-Subset Metric}
\label{sec:Subset}
While the exhaustive model benefits from a broad feature space, the EvoIQA-Subset configuration demonstrates the efficiency of Genetic Programming when restricted to a refined, high-priority feature set. This subset was constructed by employing a Random Forest (RF) regressor to rank the exhaustive feature set by Gini importance.

The GP process, restricted to this RF-prioritized space, converged on a remarkably elegant symbolic expression :

\begin{equation} Q = \Omega_{cv} \cdot \nabla_{ref} \left( \sigma_c - \sqrt{\Delta_{cs}} \cdot \Omega_{cv} \right)^2 + \Omega_{cv} 
\label{eq:evoiq-subset}\end{equation}

\begin{table}[tb]
  \caption{Definitions of the perceptual primitives in the $\mathcal{M}_{\text{EvoIQA-Subset}}$ metric.
  }
  \label{tab:headings}
  \centering
  \scriptsize
  \renewcommand{\arraystretch}{1.2} 
  \setlength{\tabcolsep}{4pt} 
  \begin{tabular}{@{}ll p{4cm}@{}}
   \toprule
    Symbol & Feature Name & Perceptual Significance\\
    \midrule
    $\Delta_{cs}$ & Chromatic-Structural Divergence & Quantifies discrepancy between chromatic and structural loss: 
$\Delta_{cs} = \mu_{S_{mn}} - \mu_{S_{gm}}$. \\

$\Omega_{cv}$ & Coeff. of Variation of structural similarity &  
Penalizes localized artifacts (e.g., JPEG blocks) over uniform noise as shown in Figure \ref{fig:four_map_comparison}. 
$\Omega_{cv} = \sigma_{S_{gm}} / \mu_{S_{gm}}$. \\

\bottomrule
\end{tabular}
\end{table}

\textit{In contrast to metrics like HaarPSI, the primary innovation of $\mathcal{M}_{\text{EvoIQA-Subset}}$ is its purely single-scale pipeline. Instead of an iterative three-level decomposition, our model operates at the base resolution using raw spatial-domain Scharr filters and LMN-channel averaging. By relying exclusively on the single-scale gradient similarity map ($s_{gm}$) and LMN chromatic similarity maps ($s_m, s_n$) parameterized using AGGD, it significantly reduces computational overhead and simplifies feature extraction.}

\begin{table}[tb]
  \caption{Performance comparison (SROCC) on the TID2013 test partition. All methods report the \textbf{mean performance across 30 independent runs}, using randomized reference-based partitioning ($20\%$ held-out test set per run). $^{\dagger}$DL baselines are cited from \cite{zhang2018blind} (comparable 10-run 80/20 protocol). $^{*}$BRISQUE was independently trained using a 10-run protocol.}
  \label{tab:comparison_results}
  \centering
   \scriptsize
  \begin{tabular}{@{}lll@{}}
    \toprule
    Method & Model Type & SROCC \\
    \midrule
    SSIM \cite{wang2004image} & Manual (Traditional) & 0.6400 \\
    MS-SSIM \cite{wang2003multiscale} & Manual (Traditional) & 0.7915 \\
    FSIM \cite{zhang2011fsim} & Manual (Traditional) & 0.8016 \\
    HaarPSI \cite{reisenhofer2018haar} & Manual (Traditional) & 0.8662 \\
    BRISQUE \cite{brisque2024}$^*$ & No-Reference (NR) & 0.5746 \\
    \midrule
    MEON$^\dagger$  & Deep Learning (Black-Box) & 0.8800 \\
    DB-CNN$^\dagger$ & Deep Learning (Black-Box) & 0.8900 \\
    \midrule
    SVR & Machine Learning (Black-Box) & 0.8131 \\
    Random Forest & Machine Learning (Ensemble) & 0.9390 \\
    \midrule
    \textbf{Evo-IQA Full (Ours)} & \textbf{Symbolic (White-Box)} & \textbf{0.8898} \\
    \bottomrule
  \end{tabular}
\end{table}

\subsection{Cross-Database Robustness}
As summarized in \cref{tab:full_tid2013_comparison} \& \cref{tab:multi_dataset_results}, we evaluated the framework through a cross-database analysis against both hand-crafted indices and deep learning architectures. Key findings include:

\begin{itemize}
    \item \textbf{Efficiency and Superiority over Deep Learning:} When trained on KADID-10k and tested on TID2013, \textit{EvoIQA-Full} achieves a peak SRCC of $0.8979$. This significantly outperforms the HaarPSI baseline ($0.870$), as well as specialized deep networks trained on KADID-10k and tested on TID2013 such as DBCNN ($0.689$) \cite{zhang2018blind}, CoDI-IQA($0.786$) \cite{liu2025content} with the performance metrics for these DNN baselines cited directly from their reported results in \cite{liu2025content}.Our model achieves this comparable accuracy without the massive computational burden of a 47-million-parameter architecture. 

    \item \textbf{Generalization \& Efficiency:} Despite exclusive training on TID2013, the models generalize to unseen datasets (CID:IQ, VDID), surpassing SVR and Random Forest benchmarks while using fewer than 10 parameters—yielding a transparent, explainable model.
    
    \item \textbf{Statistical Stability:} Across 30 independent runs, the metrics exhibit negligible variance ($\pm 0.006$), confirming the consistent convergence of the Genetic Programming process compared to the stochasticity often observed in deep learning training \cite{picard2021torch, ehrhardt2022critical}.
\end{itemize}

\begin{table*}[t]
  \caption{Cross-dataset SROCC Performance Comparison. All models were trained on the TID2013 reference dataset to evaluate generalization capability. For EvoIQA, we report the \textbf{Mean $\pm$ Standard Deviation} across 30 runs, with the \textbf{Maximum Absolute SROCC} achieved shown in brackets. Both Random Forest and SVR were trained on the feature subset used by Evo-IQA$_{\text{Subset}}$.}
  \label{tab:cross_data}
  \centering
  \scriptsize
  \begin{tabular}{@{}lcccc@{}}
    \toprule
    \textbf{Method} & \textbf{KADID-10k} & \textbf{CSIQ} & \textbf{CID:IQ} & \textbf{VDID} \\
    \midrule
    SS-SSIM       & 0.720 & 0.864 & 0.773 & 0.895 \\
    MS-SSIM       & 0.800 & 0.913 & 0.793 & 0.901 \\
    FSIM          & 0.830 & 0.924 & 0.779   & 0.926   \\
    HaarPSI       & \textbf{0.890} & \textbf{0.956} & 0.755 & 0.920 \\
    Random Forest & 0.886 & 0.894 & 0.670 & 0.904 \\
    SVR           & 0.778 & 0.880 & 0.511 & 0.857 \\
    \midrule
    \textit{Our Proposed Metrics} \\
    \textbf{EvoIQA-Full}  & 0.866 $\pm$ 0.008 & 0.914 $\pm$ 0.011 & 0.750 $\pm$ 0.020 & 0.910 $\pm$ 0.005 \\
                         & [0.880]           & [0.930]           & \textbf{[0.803]}  & [0.915]           \\
    \textbf{EvoIQA-Subset} & 0.864 $\pm$ 0.001 & 0.912 $\pm$ 0.014 & 0.700 $\pm$ 0.012 & \textbf{0.912 $\pm$ 0.008} \\
                         & [0.875]           & [0.944]           & [0.721]           & \textbf{[0.931]}  \\
    \bottomrule
  \end{tabular}
\end{table*}

\begin{table*}[b!]
  \caption{Performance comparison (SROCC) across all 24 TID2013 test partition distortion categories. Results represent the mean across 30 independent runs. Rows highlighted in \colorbox{green!15}{green} indicate categories where \textbf{EvoIQA$_{\text{Full}}$}  achieves the highest correlation. Deep learning baselines (MEON, DB-CNN) are cited from \cite{zhang2018blind}. The results are statistically significant with  ($p < 0.05$).}
  \label{tab:full_tid2013_comparison}
  \centering
   \scriptsize
  \begin{tabular}{@{}lccc|c@{}}
    \toprule
    \textbf{Distortion Category} & \textbf{MEON} & \textbf{DB-CNN } & \textbf{HaarPSI} & \textbf{EvoIQA$_{\text{Full}}$ (Ours)} \\ 
    \midrule
    \rowcolor{green!15} Additive Gaussian noise & 0.813 & 0.790 & 0.943 & \textbf{0.952} \\
    \rowcolor{green!15} Additive noise in color components & 0.722 & 0.700 & 0.867 & \textbf{0.886} \\
    \rowcolor{green!15} Spatially correlated noise & 0.926 & 0.826 & 0.941 & \textbf{0.943} \\
    Masked noise & 0.728 & 0.646 & \textbf{0.773} & 0.750 \\
    \rowcolor{green!15} High frequency noise & 0.911 & 0.879 & 0.913 & \textbf{0.928} \\
    Impulse noise & \textbf{0.901} & 0.708 & 0.885 & 0.870 \\
    \rowcolor{green!15} Quantization noise & 0.888 & 0.825 & 0.904 & \textbf{0.910} \\
    \rowcolor{green!15} Gaussian blur & 0.887 & 0.859 & 0.920 & \textbf{0.948} \\
    \rowcolor{green!15} Image denoising & 0.797 & 0.865 & 0.942 & \textbf{0.953} \\
    \rowcolor{green!15} JPEG compression & 0.850 & 0.894 & 0.954 & \textbf{0.956} \\
    JPEG2000 compression & 0.891 & 0.916 & \textbf{0.970} & 0.960 \\
    JPEG transmission errors & 0.746 & 0.772 & \textbf{0.886} & 0.870 \\
    \rowcolor{green!15} JPEG2000 transmission errors & 0.716 & 0.773 & 0.924 & \textbf{0.926} \\
    Non-eccentricity pattern noise & 0.116 & 0.270 & \textbf{0.808} & 0.794 \\
    \rowcolor{green!15} Local block-wise distortions & 0.500 & 0.444 & 0.482 & \textbf{0.656} \\
    \rowcolor{green!15} Mean shift (intensity shift) & 0.177 & -0.009 & 0.731 & \textbf{0.805} \\
    \rowcolor{green!15} Contrast change & 0.252 & 0.548 & 0.435 & \textbf{0.466} \\
    \rowcolor{green!15} Change of color saturation & 0.684 & 0.631 & 0.710 & \textbf{0.803} \\
    \rowcolor{green!15} Multiplicative Gaussian noise & 0.849 & 0.711 & 0.895 & \textbf{0.912} \\
    \rowcolor{green!15} Comfort noise & 0.406 & 0.752 & 0.927 & \textbf{0.928} \\
    Lossy compression of noisy images & 0.772 & 0.860 & \textbf{0.960} & 0.956 \\
    \rowcolor{green!15} Color quantization with dither & 0.857 & 0.833 & 0.903 & \textbf{0.929} \\
    \rowcolor{green!15} Chromatic aberrations & 0.779 & 0.732 & 0.833 & \textbf{0.850} \\
    \rowcolor{green!15} Sparse sampling and reconstruction & 0.855 & 0.902 & 0.961 & \textbf{0.963} \\
    \bottomrule
  \end{tabular}
\end{table*}

\subsection{Performance by distortion groups}
\label{sec:dist_groups}
For this subsection, we will limit the discussion to the Evo-IQA Full model. The performance of the Evo-IQA model across TID2013, KADID-10k, and CID-IQ as reported in \cref{tab:full_tid2013_comparison} \& \cref{tab:multi_dataset_results}  demonstrates robust generalization, outperforming HaarPSI in 18 of 24 distortion types on the TID2013 dataset. Significantly, our symbolic framework frequently surpasses high-performance Deep Neural Network (DNN) architectures, including MEON \cite{ma2018end} and DB-CNN \cite{zhang2018blind}. It is important to note that these results are from their original publications, which were validated using an 80/20 train-test split on TID2013 over 10 independent runs. 
The most significant improvements occur in specialized categories, notably Local block-wise distortions (Evo-IQA $0.6557$ vs. HaarPSI's $0.4822$) and Change of color saturation (Evo-IQA $0.8034$ vs. $0.7100$). Even in traditional categories like JPEG Compression and Gaussian Blur, the model maintains a superior SROCC.

\begin{table}[tb]
  \caption{Cross-dataset SROCC Performance Comparison. Results are reported as the Mean [Maximum] performance across 30 independent runs. Rows highlighted in \colorbox{green!15}{green} indicate categories where the \textbf{EvoIQA$_{\text{Full}}$} outperforms the HaarPSI baseline; \textbf{bolded} values indicate where the \textbf{EvoIQA$_{\text{Full}}$ Max} exceeds the baseline.}
  \label{tab:multi_dataset_results}
    \scriptsize
  \centering
  \begin{tabular}{@{}llcc@{}}
    \toprule
    \textbf{Dataset} & \textbf{Category} & \textbf{HaarPSI} & \textbf{EvoIQA$_{\text{Full}}$ (Mean [Max])} \\ 
    \midrule
    KADID-10k & Blurs               & 0.9389 & 0.9223 [0.9354] \\
              & Color Distortions   & 0.7875 & \textbf{0.7609 [0.8107]} \\
              & Compression         & 0.9174 & 0.8824 [0.9003] \\
              & Noise               & 0.9181 & 0.8964 [0.9115] \\
              & Brightness Change   & 0.8406 & 0.8217 [0.8396] \\
              & Spatial Distortions & 0.6662 & 0.5975 [0.6359] \\
              & Sharpness/Contrast  & 0.8431 & 0.8028 [0.8361] \\ 
    \midrule
    CID-IQ    & DeltaE Gamut Map    & 0.9307 & 0.9058 [0.9208] \\
              & \\rowcolor{green!15} SGCK Gamut Map & 0.8383 & \textbf{0.8384 [0.8751]} \\
              & Poisson Noise       & 0.8212 & \textbf{0.7859 [0.8292]} \\
              & JPEG2000 Comp.      & 0.7597 & \textbf{0.7500 [0.7805]} \\
              & Gaussian Blur       & 0.7680 & 0.6935 [0.7452] \\
              & JPEG Compression    & 0.7679 & \textbf{0.6867 [0.8090]} \\ 
    \bottomrule
  \end{tabular}
\end{table}

\begin{table}[tb]
\centering
\caption{Ablation Study of Hyper-parameters on TID2013\textsubscript{test}}
\label{tab:ablation}
  \centering
    \scriptsize
    
  \begin{tabular}{@{}lll@{}}
    \toprule
    \textbf{Configuration} & \textbf{SROCC} & \textbf{$\Delta$ SROCC} \\ 
    \midrule
    EvoIQA$_{\text{Full}}$  & \textbf{0.8898}  & --- \\
    EvoIQA$_{\text{Full}}$ (Pearson Correlation Fitness) & 0.8863  & -0.0035 \\
    Non-Linear Operators ($\log, \exp$) & 0.8894  & -0.0004 \\
    Reduced Population ($N=200$) & 0.8825 & -0.0073 \\
    Cold-Start (No Init. \cref{subsec:evolution_process}, $N=200$) & 0.8810  & -0.0088 \\
    \bottomrule
  \end{tabular}
\end{table}

\section{Conclusions}
\label{sec:conclusion}
In this study, we successfully developed an interpretable Image Quality Assessment framework using Genetic Programming. By evolving symbolic formulas from high-level perceptual features—including Phase Congruency, Gradient Magnitude, and Chromaticity Similarity — our framework achieves performance parity with, and often exceeds, state-of-the-art "black-box" DNNs like MEON and DB-CNN.  Rigorous 30-run validations confirm that the evolutionary process is robust to different data partitions yielding robust, lightweight models with low predictive variance. This work demonstrates that symbolic AI can provide "White-Box" IQA solutions without the computational overhead of deep learning. Lastly, we aim to evolve Reference-Free metrics by applying our Genetic Programming engine directly to the AGGD parameters of distorted images. This paradigm shift towards NR-IQA would leverage Natural Scene Statistics (NSS) to assess perceptual quality without requiring access to the original, undistorted signal.

\clearpage  

%
%
\bibliographystyle{splncs04}
\bibliography{main}
\end{document}